# DEVELOPING PARALLEL DEPENDENCY GRAPH IN IMPROVING GAME BALANCING

Sim Hui Tee
*Multimedia University*
*Jalan Multimedia 63100 Cyberjaya,*
*Selangor, Malaysia*

**ABSTRACT**

The dependency graph is a data architecture that models all the dependencies between the different types of assets in the game. It depicts the dependency-based relationships between the assets of a game. For example, a player must construct an arsenal before he can build weapons. It is vital that the dependency graph of a game is designed logically to ensure a logical sequence of game play. However, a mere logical dependency graph is not sufficient in sustaining the players' enduring interests in a game, which brings the problem of game balancing into picture. The issue of game balancing arises when the players do not feel the chances of winning the game over their AI opponents who are more skillful in the game play. At the current state of research, the architecture of dependency graph is monolithic for the players. The sequence of asset possession is always foreseeable because there is only a single dependency graph. Game balancing is impossible when the assets of AI players are overwhelmingly outnumbering that of human players. This paper proposes a parallel architecture of dependency graph for the AI players and human players. Instead of having a single dependency graph, a parallel architecture is proposed where the dependency graph of AI player is adjustable with that of human player using a support dependency as a game balancing mechanism. This paper exhibits that the parallel dependency graph helps to improve game balancing.

## 1. INTRODUCTION

This paper aims at improving game balancing using parallel dependency graph in modeling all the dependency relationship between the different types of assets in the game. The dependency graph is a data architecture that defines the sequence of asset possession of a game player. For example, a player must construct an arsenal before he can build a light panzer. He must have five light panzers before he can construct a battle tank, and so on. The human player familiarizes with the sequence of asset possession after spending some time in the game play because there is a fixed single stream of dependency relationship between assets. The advantage of this single dependency graph architecture is that it promotes consistency. However, the major disadvantage of this architecture is that it is incapable of sustaining the human players' interest to complete the game when there is no hope of winning. For instance, when the tanks of AI opponent have destroyed all arsenals and economic resources of the human player, the overwhelming outcome leads to no hope of victory for human player. The human player will lose the enthusiasm to continue the game towards the end. This issue is known as game balancing (Björk and Holopainen 2005). The main challenge of striking game balancing in any genre of games is to maintain an adjustable balance of capabilities between the players. A game that always achieves game balancing tends to draw and sustain the players' interest because they believe there are chances of victory. Game balancing plays an important role in creating a meaningful play in a game.

There are two types of dependency graph. The primary dependency type is a creational dependency (Tozour 2001) where the sequence of existence of each asset/entity is defined rigorously. For instance, the base must be created prior to the school, while the school should be built before a college can be constructed. Apart from including these entity dependencies, creational dependency also includes abstract dependencies such as resource dependency and social rank dependency in a game. Creational dependency maintains a consistent outlook of a game. However, it is not flexible in response to the potentiality of game imbalance. For example, Figure 1 illustrates a game where a player needs to go through certain stages before acquiring an advanced asset. When the fundamental asset, which is the pre-requisite of other assets, has been destructed by the AI player, imbalance of game is highly probable. In Figure 1, bank is the pre-requisite asset before a university, barrack and armory can be created. If the AI player has destroyed the banks, imbalance of game is the possible consequence as the human player may not have sufficient time to acquire advanced weapons to fight with AI player. The human player is forced to abide to the creational dependency path without skipping any pre-requisite asset acquisition.

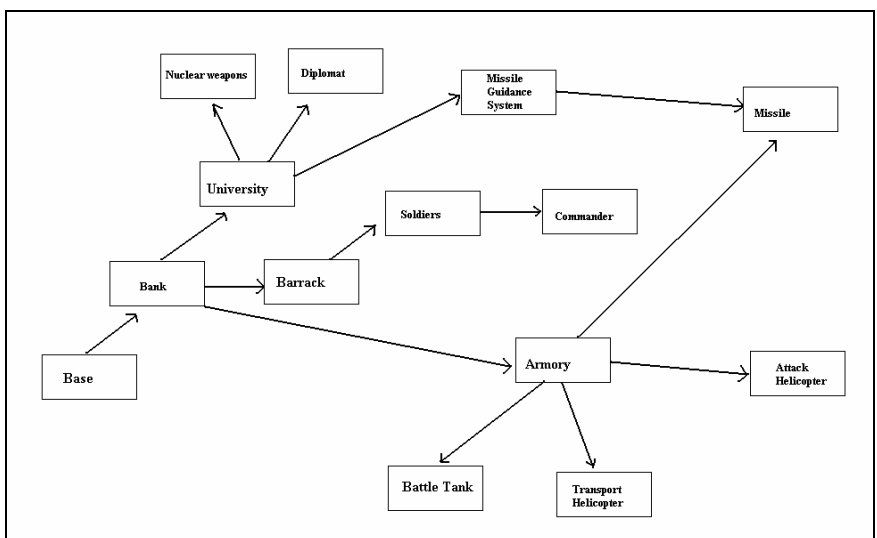

Figure 1: Creational dependency graph of a battle game.

The secondary dependency type is a support dependency (Tozour 2001). It is used to support the creational dependency. In this research, support dependency is used in two levels. Firstly, it is a dependency that supports the function of a particular game entity in the creational dependency graph. Secondly, support dependency plays the role as a bridging dependency which coordinates more than one creational dependency graph. It is the second aspect of support dependency that developed as a novel approach in this paper to realize the parallel creational dependency graph for the purpose of improving game balancing.

## 2. SUPPORT DEPENDENCY FOR PARALLEL DEPENDENCY GRAPH

A dependency graph is useful in evaluating the competency of game players. In this research, each parent dependency node is assigned a set of relative values, namely "dependency value", corresponding to its child nodes. Dependency values indicate the significance of the parent node to the child node in term of winning the control over the game. The higher the dependency value, the more vital a parent node to the child node in term of game resources. Similarly, the lower dependency value implies the shorter time and lesser resources for the child node creation. Based on the dependency graph in Figure 1, the example of dependency values is given in Table 1 below.

Table 1: Dependency values

| | Bank | Barrack | Uni | N. weapons | Diplomat | Missile GS | Missile | Soldiers | Commander | Armory | A. Helicopter | T. Helicopter | Battle Tank |
|---|---|---|---|---|---|---|---|---|---|---|---|---|---|
| Base | 1 | 2 | 3 | 5 | 5 | 5 | 7 | 4 | 5 | 3 | 5 | 5 | 5 |
| Bank | | 1 | 1 | 4 | 4 | 4 | 6 | 3 | 4 | 1 | 3 | 3 | 3 |
| Barrack | | | | | | | | 1 | 2 | | | | |
| University | | | | 2 | 2 | 2 | 3 | | | | | | |
| N.Weapons | | | | | | | | | | | | | |
| Diplomat | | | | | | | | | | | | | |
| Missile GS | | | | | | | 1 | | | | | | |
| Missile | | | | | | | | | | | | | |
| Soldiers | | | | | | | | | 1 | | | | |
| Commander | | | | | | | | | | | | | |
| Armory | | | | | | | 2 | | | | 2 | 2 | 2 |
| A.Helicopter | | | | | | | | | | | | | |
| T.Helicopter | | | | | | | | | | | | | |
| Battle Tank | | | | | | | | | | | | | |

In Table 1, the dependency value of Base-Soldiers is lower than that of Base-Missile. This implies that the base is more significant for the creation of missile. If the base has been destroyed, the player needs more time and resources to create the high-tech weapon such as missile. The imbalance of game is likely to happen when the AI player has plenty of high-tech weapons while the human player has not.

To restore the balance of the game, it is necessary to calculate the aggregate path dependency value along each dependency path, as shown in Table 2. The aggregate values derived from the sum of dependency values in each row in Table 1.

Table 2: Aggregate path dependency value along each dependency path

| Nodes/assets | Aggregate path dependency values |
|---|---|
| Base | 55 |
| Bank | 37 |
| Barrack | 3 |
| University | 9 |
| N.weapons | 0 |
| Diplomat | 0 |
| Missile GS | 1 |
| Missile | 0 |
| Soldiers | 1 |
| Commander | 0 |
| Armory | 8 |
| A.Helicopter | 0 |
| T.Helicopter | 0 |
| Battle tank | 0 |

The higher the aggregate path dependency value, the more significant an asset in a game. In Table 2, Base has the highest aggregate path dependency value, followed by Bank and University. This indicates the three most valuable game assets are Base, Bank and University, respectively. The respective cease-fire time is then assigned to the AI player after the devastation of each human player's asset, as demonstrated in Table 3. The cease-fire time is mapped with the aggregate path dependency value.

Table 3: Imposing cease-fire time on AI player

| Destroyed assets of human player | Cease-fire time imposed on AI player (second) |
|---|---|
| Base | 900 |
| Bank | 550 |
| Barrack | 250 |
| University | 400 |
| N.weapons | 0 |
| Diplomat | 0 |
| Missile GS | 120 |
| Missile | 0 |
| Soldiers | 120 |
| Commander | 0 |
| Armory | 380 |
| A.Helicopter | 0 |
| T.Helicopter | 0 |
| Battle tank | 0 |

The purpose of imposing cease-fire time on AI player is to allow the human player sufficient time to rebuild his assets. Longer cease time will be imposed on AI player if it has destroyed an asset of human player which has higher aggregate path dependency value. This method helps to restore the game balancing.

The No-Attack instruction can be built into the game balancing mechanism that is to be invoked when the assets of human player has been destroyed. The game balancing mechanism serves as a support dependency that bridges and balances the dependency graphs of human and AI players. Figure 2 illustrates the parallel dependency graph which is bridged by the balancing mechanism (support dependency) when the base of human player has been destructed. The balancing mechanism will be invoked and imposing the No-Attack instruction on AI player for 900 seconds.

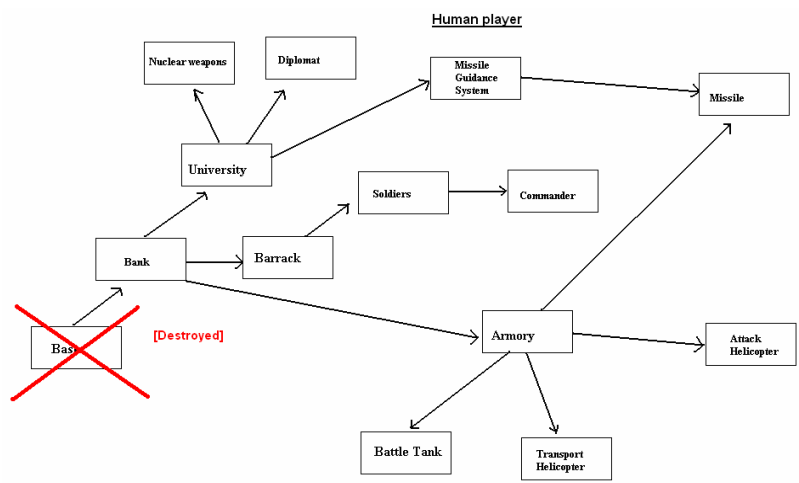

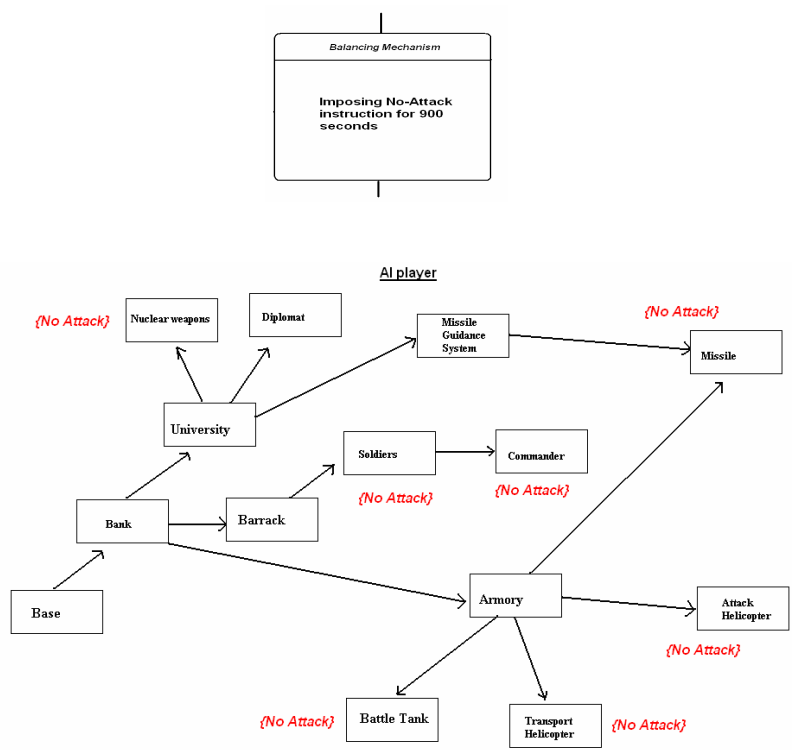

Figure 2: Balancing mechanism as a support dependency to balance the parallel dependency graphs between human and AI players

## 3. CONCLUSION

Meaningful play in a game is largely depending on the relationship between player action and game outcome (Salen and Zimmerman 2004). Freeman (2004) holds that balancing the degree and ways in which the player can affect the game, or at least seem to affect it, contribute to producing a meaningful and balanced game. This research attempts to demonstrate that parallel dependency graph can be used as an approach in achieving game balancing in any genre of game development. The advantage of parallel dependency graph is the adjustability of players' competency level. It minimizes the potential overwhelming victory on AI side in order to sustain human player's enthusiasm in the game. Parallel dependency graph can be used as an approach in developing games for any platform with the objective of maximizing game balancing.

## REFERENCES

Björk, S and Holopainen, J, 2005. *Patterns in Game Design.* Charles River Media, USA.

Freeman, D. 2004. *Creating Emotion in Games*. New Riders Publishing, USA.

Salen, K and Zimmerman, E, 2004. Rules of Play: *Game Design Fundamentals*. The MIT Press, USA.

Tozour, P. 2001. Strategic Assessment Techniques. In: DeLoura, M, ed. 2001. *Game Programming Gems 2*. Charles River Media. Pp 298-306.